\documentclass[runningheads]{llncs}
\usepackage{cite}
\usepackage{amsmath,amssymb,amsfonts}
\usepackage{algorithmic}
\usepackage{graphicx}
\usepackage{textcomp}
\usepackage{xcolor}
\usepackage{url}
\usepackage{multirow}
\usepackage{booktabs}
\usepackage{verbatim}
\usepackage{bbm}
\usepackage{marvosym}
\begin{document}
\title{Hypergraph Convolutional Networks for Fine-grained ICU Patient Similarity Analysis and Risk Prediction}
%
%
\author{Yuxi Liu\inst{1}\textsuperscript{(\Letter)} \and Zhenhao Zhang\inst{2} \and Shaowen Qin\inst{1} \and Flora D. Salim\inst{3} \and Antonio Jimeno Yepes\inst{4} \and Jun Shen\inst{5} \and Jiang Bian\inst{6}}
\institute{College of Science and Engineering, Flinders University, Tonsley, SA 5042, Australia
\email{\{liu1356, shaowen.qin\}@flinders.edu.au} \\ \and
College of Life Sciences, Northwest A\&F University, Yangling, Shaanxi 712100, China
\email{\{zhangzhenhow\}@nwafu.edu.cn} \\ \and
School of Computer Science and Engineering, UNSW, Sydney, NSW 2052, Australia
\email{\{flora.salim\}@unsw.edu.au} \\ \and
School of Computing Technologies, RMIT University, Melbourne, VIC 3001, Australia
\email{\{antonio.jose.jimeno.yepes\}@rmit.edu.au} \\ \and
School of Computing and Information Technology, UOW, Wollongong, NSW 2522, Australia
\email{\{jshen\}@uow.edu.au} \\ \and
College of Medicine, University of Florida, Gainesville, FL, 32610, USA
\email{\{bianjiang\}@ufl.edu} \\
}

%
%
%
\pagestyle{empty}
\thispagestyle{empty}
\maketitle              
\begin{abstract}
The Intensive Care Unit (ICU) is one of the most important parts of a hospital, which admits critically ill patients and provides continuous monitoring and treatment. Various patient outcome prediction methods have been attempted to assist healthcare professionals in clinical decision-making. Existing methods focus on measuring the similarity between patients using deep neural networks to capture the hidden feature structures. However, the higher-order relationships are ignored, such as patient characteristics (e.g., diagnosis codes) and their causal effects on downstream clinical predictions. In this paper, we propose a novel Hypergraph Convolutional Network that allows the representation of non-pairwise relationships among diagnosis codes in a hypergraph to capture the hidden feature structures so that fine-grained patient similarity can be calculated for personalized mortality risk prediction. Evaluation using a publicly available eICU Collaborative Research Database indicates that our method achieves superior performance over the state-of-the-art models on mortality risk prediction. Moreover, the results of several case studies demonstrated the effectiveness and robustness of the model decisions.

\keywords{Hypergraph Learning \and Patient Similarity \and Electronic Health Records \and Intensive Care Unit.}
\end{abstract}

\section{Introduction}
Patient similarity can be estimated by calculating the distance between the vectors representing patient characteristics (e.g., diseases and medical events) \cite{brown2016patient}. In clinical practice, healthcare professionals often use specific disease characteristics and adverse events to group similar patients to facilitate predictive clinical and managerial decision-making \cite{lee2015personalized}. However, due to variations in patient conditions and treatment needs, more robust methods need to be developed and introduced to support decisions, as illustrated by \cite{wollenstein2020personalized, mcrae2020managing, heo2021prediction} on the need for personalized patient care in ICU during the COVID-19 pandemic.

Machine learning has great potential to guide clinical practice \cite{adlung2021machine, johnson2021precision}. For instance, machine learning-based prediction models can be used for a wide range of clinical applications, including predicting the risk of in-hospital mortality and physiologic decline, estimating hospital length of stay, and classifying phenotype \cite{harutyunyan2019multitask, sheikhalishahi2020benchmarking}. When used on larger medical datasets, traditional machine learning-based prediction models are limited by their simple architecture design, making personalized prediction extremely challenging. Deep personalized prediction models based on similar patients have emerged \cite{zhu2016measuring, ni2017fine, suo2018deep} to deal with the challenge. These models are trained using information from similar patients and thus have the potential to identify risk factors associated with individual patients.

Increased application of Graph Neural Networks (GNNs) has been observed across a number of research fields in recent years, such as social recommender systems \cite{fan2019graph}, bioinformatics \cite{zhang2021graph}, and knowledge graphs \cite{wu2020temp}. GNN is a type of neural network for dealing with graph-structured data \cite{zhang2019learning, dash2021incorporating}. When used on high-dimensional or complex data, very deep GNNs can be constructed by stacking multiple graph convolutional layers. In GNNs, each graph's convolutional layer aggregates information from neighboring nodes and edges using a message-passing strategy. At each GNN message-passing iteration, each node aggregates information from its neighborhood, and as these iterations progress, each node embedding reaches out further in the graph to extract global information. By doing so, both local and global information from the graph is taken into consideration for generating useful node and graph-level representations for various downstream predictions.

Most research on GNNs has been carried out on pairwise relationships of objects of interest \cite{wu2021enhancing, he2022gnnrank, liu2022pilsl}. Previous studies by \cite{liu2020heterogeneous, wang2020graph, gu2022structure} have demonstrated the effectiveness of GNN-based models on patient similarity computation. These studies mainly focus on using GNNs to learn the representation of pairwise interaction between two patients for downstream applications. However, in many real medical applications, the relationships between patient characteristics (e.g., diagnosis codes) are beyond what a pairwise formulation can represent.

Recently, hypergraphs have been utilized for the modeling of a wide range of systems where high-order relationships exist among their interacting parts. Current studies have demonstrated the effectiveness of hypergraphs in encoding high-order data correlation \cite{feng2019hypergraph, yang2022efficient, cai2022hypergraph}. Motivated by these successful applications, in this paper, we aim to capture the non-pairwise relationships among patient characteristics by modeling structured electronic health record (EHR) data$\footnote{EHRs are longitudinal electronic records of patient health information including both structured (e.g., vital signs, lab tests, and diagnosis codes) and unstructured (e.g., free-text reports and X-ray images) data \cite{williams2017clinical}.}$ with the utilization of a Hypergraph Convolutional Network (Figure~\ref{fig:OVERVIEW}). The intuition behind introducing Hypergraph Convolutional Network can be explained as seeing the need to specify prior medical knowledge (i.e., previous ICD-9 diagnosis codes) that is in the form of higher-order relationships in a hypergraph. Based on the foundation established by hypergraph representation learning, we conduct patient similarity computation and then aggregate the information from similar patients as we analyze patient graphs. To demonstrate the efficacy of the proposed method, we conduct the experiments on predicting in-hospital mortality risks for ICU patients using the publicly available eICU Collaborative Research Database \cite{pollard2018eicu}.

Our contributions are as follows:
\begin{itemize}
    \item We introduce Hypergraph Convolutional Network for fine-grained ICU patient similarity computation. To the best of our knowledge, this is the first study that uses a tailored Hypergraph Convolutional Network for similarity computation among patients in ICU settings.
    \item We evaluate our method against deep prediction methods on the publicly available eICU Collaborative Research Database, and the results surpass the state-of-the-art models in AUROC, AUPRC, Precision, F1 Score, and Min(Se, P+). Moreover, we demonstrate the advantages of our method in terms of effectiveness and robustness of decisions with several case studies.
\end{itemize}

\section{Related Work}
A considerable literature has been published around the theme of EHR-based risk prediction \cite{ma2018risk, ye2020lsan, luo2020hitanet, zhang2020inprem, ma2020adacare, cui2022automed}. Representative ICU mortality risk prediction models include \cite{pirracchio2015mortality, awad2017early, ge2018interpretable, che2018recurrent}. It is worth mentioning that the traditional SAPS \cite{le1984simplified} and APACHE \cite{knaus1981apache} scores, as well as their variants SAPS II \cite{le1993new} and APACHE II \cite{knaus1985apache} scores, are mainly used for assessing the severity of the health condition as defined by the probability of patient mortality.

The study by \cite{pirracchio2015mortality} introduced the Super Learner Algorithm (SICULA) to predict mortality risk for ICU patients. The SICULA is an ensemble machine-learning framework that comprises a series of traditional machine-learning models, such as generalized linear models. Experimental results on the MIMIC-II dataset demonstrate that SICULA outperforms the traditional SAPS-II and APACHE-II scores. Similarly, \cite{awad2017early} proposed an ensemble machine learning framework (EMPICU Random Forest) to predict the mortality risk of patients based on data from the first 24 hours and 48 hours after ICU admission. Experimental results on the MIMIC-II dataset demonstrate that EMPICU Random Forest outperforms the traditional SAPS-I and APACHE-II scores, random forests, decision trees, etc. The ICU-LSTM \cite{ge2018interpretable} was proposed to take both sequential and non-sequential features as inputs for ICU mortality risk prediction. The former refers to vital signs, while the latter refers to the previous ICD-10 diagnosis codes. ICU-LSTM is built with the Long short-term memory (LSTM) units \cite{hochreiter1997long}. Experimental results on the Asan Medical Center (AMC) ICU dataset demonstrate that ICU-LSTM outperforms the traditional logistic regression model. The study by \cite{che2018recurrent} proposed GRU-D to model the long-term temporal dependencies in multivariate clinical time series and utilize the decay mechanism to learn the impact of varying time intervals. GRU-D \cite{che2018recurrent} is built upon Gated Recurrent Unit (GRU) \cite{cho2014learning}. The GRU is a variant of recurrent neural networks featured with a reset gate and an update gate, which control the flow of information between the hidden state and the current input. Experimental results on the MIMIC-III and PhysioNet datasets demonstrate that GRU-D achieves superior performance over the state-of-the-art models on mortality risk prediction.

\section{Method}
\begin{figure*}[!htbp]
        \centering
        \includegraphics[width = 1.0\linewidth]{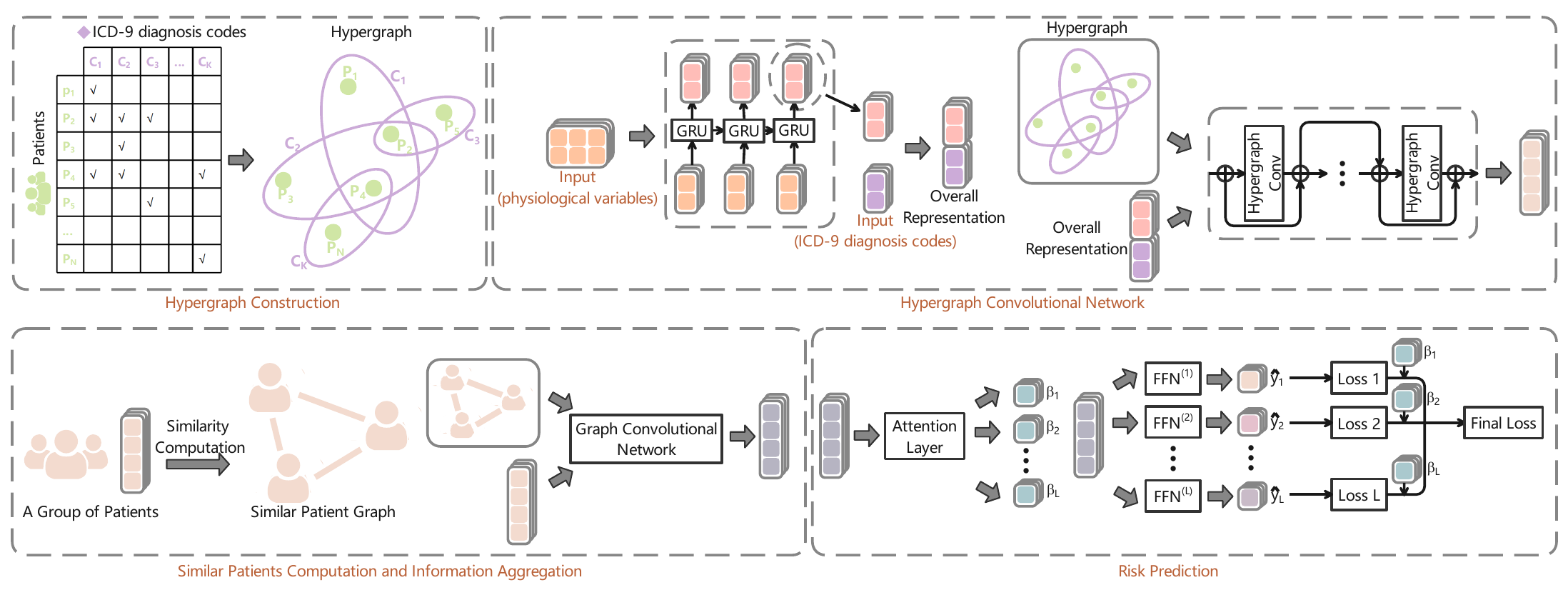}
        \caption{\textbf{Schematic description of the proposed method.}}
        \label{fig:OVERVIEW}
\end{figure*}

In this section, we design Hypergraph Convolutional Network (Figure~\ref{fig:OVERVIEW}) to specify prior medical knowledge in the form of higher-order relationships in a hypergraph. Based on the foundation established by hypergraph representation learning, we conduct patient similarity computation and then aggregate the information from similar patients as we analyze patient graphs.

\subsection{Data Representation}
EHR data contains patients' time-ordered records. Each patient's records ensemble is a multivariate clinical time series with up to $M$ physiological variables, denoted by $X = \{X_{1}, X_{2}, \cdots, X_{T}\} \in \mathbb{R}^{M \times T}$, where $T$ is the total number of records. For instance, $X_{t} = \{X_{t}^{1}, X_{t}^{2}, \cdots, X_{t}^{M}\} \in \mathbb{R}^{M}$ is the t-th record and $X_{t}^{m}$ is the value of the m-th physiological variable of $X_{t}$. EHR data also contains ICD-9 diagnosis codes, denoted by $X^{ICD} \in \{0,1\}^{g}$.

\subsection{Hypergraph Convolutional Network}
To construct EHR data into a hypergraph, we regard each ICD-9 diagnosis code as a hyperedge and each sample/patient as a node, where each hyperedge connects all input samples/patients who have the same ICD-9 diagnosis codes. We define the hypergraph as $G$ = $\{V, E\}$, where $V$ and $E$ are sets of $N$ nodes and $K$ hyperedges. Each hyperedge $e \in E$ has a positive weight $W_{ee}$, with all the positive weights placed on a diagonal matrix $W \in \mathbb{R}^{K \times K}$. Accordingly, the hypergraph $G$ can be denoted by an incidence matrix $H \in \{0,1\}^{N \times K}$. When the hyperedge $e \in E$ is incident with a node $v_{i} \in V$, i.e., $v_{i}$ is connected with $e$, $H_{ie} = 1$, otherwise 0.

Now we feed each patient's records ensemble into GRUs \cite{cho2014learning} to generate a series of hidden representations, i.e., $\bar{X}_{1}, \bar{X}_{2}, \cdots, \bar{X}_{T} = GRU(X_{1}, X_{2}, \cdots, X_{T})$. We then take the hidden representation at time $T$ (i.e., $\bar{X}_{T} \in \mathbb{R}^{d}$, note that we drop the subscript $T$ in the following steps), and concatenate $\bar{X}$ with $X^{ICD}$ as $\tilde{X} \in \mathbb{R}^{d + g}$.

Given the overall representation $\tilde{X}$, we apply the convolution operation to the hypergraph $G$ as:
\begin{equation}
\begin{split}
\label{eq:2}
\tilde{X}^{(1)} = \sigma (D^{-1} HWB^{-1}H^{\top}\tilde{X}\Theta^{(0)}),
\end{split}
\end{equation}
where $H \in \mathbb{R}^{N \times K}$ is the incidence matrix, $W \in \mathbb{R}^{K \times K}$ is the diagonal hyperedge weight matrix, $D \in \mathbb{R}^{N \times N}$ and $B \in \mathbb{R}^{M \times N}$ are degree matrices (i.e., $D_{ii} = \sum_{e = 1}^{K} W_{ee} \cdot H_{ie}$ and $B_{ee} = \sum_{i = 1}^{N} H_{ie}$), $\Theta^{(0)}$ is a layer-specific learnable parameter, and $\sigma (\cdot)$ is an activation function.

Through the processes above, we have been able to capture the information of nodes from the first-order neighborhood (i.e., $\tilde{X}^{(1)}$). To further capture the information of nodes from higher-order neighborhoods, multiple hypergraph convolutional layers are developed and introduced into the network architecture as:
\begin{equation}
\begin{split}
\label{eq:3}
\tilde{X}^{(l)} = \sigma(D^{-1}HWB^{-1}H^{\top}\tilde{X}^{(l - 1)}\Theta^{(l - 1)}),
\end{split}
\end{equation}
where $l$ is the total number of layers.
Moreover, we forward an identity mapping of each hypergraph convolutional layer's input to its output side. Accordingly, the corresponding residual block is only required to capture the difference between input and output \cite{he2016deep}, which in turn allows for building very deep neural networks and avoiding exploding/vanishing gradients.

\subsection{Similar Patients Computation and Information Aggregation}
We conduct similarity computation among the patients and then aggregate the information from similar patients. Given the input $\tilde{X}^{(l)}$, the pairwise similarities that correspond to any two patient representations as:
\begin{equation}
\begin{split}
\label{eq:4}
A = sim(\tilde{X}^{(l)}, \tilde{X}^{(l)}) = \frac{\tilde{X}^{(l)} \cdot \tilde{X}^{(l)\top}}{(d + g)^{2}},
\end{split}
\end{equation}
where $sim(\cdot)$ is the measure of cosine similarity. Subsequently, a learnable threshold $\zeta$ is introduced to $A$ as:
\begin{equation}
\begin{split}
\label{eq:5}
A^{\prime} =
  \begin{cases}
    1,  & if\ A > \zeta \\
    0,  & otherwise
  \end{cases},
\end{split}
\end{equation}
where $A^{\prime}$ is the adjacency matrix and similarities above the threshold $\zeta$ are preserved.

We treat the patients' representations as a graph to aggregate the information from similar patients as:
 \begin{equation}
\begin{split}
\label{eq:6}
X^{*} = \sigma(\tilde{D}^{-\frac{1}{2}} \tilde{A} \tilde{D}^{-\frac{1}{2}} \tilde{X}^{(l)} \Phi),
\end{split}
\end{equation}
where $\tilde{A} = A^{\prime} + I$ is the adjacency matrix with inserted self-loops, and $I$ is the identity matrix, $\tilde{D}$ is the degree matrix, i.e., $\tilde{D}_{ii} = \sum_{j}\tilde{A}_{ij}$, $\Phi$ is a learnable parameter, $\sigma(\cdot)$ is an activation function.

\subsection{Risk Prediction}
We feed $X^{*}$ into a feed-forward neural network (FFN) ensemble to make risk predictions. We take the i-th FNN as an example:
 \begin{equation}
\begin{split}
\label{eq:7}
\hat{y}^{(i)} = FFN^{(i)} (X^{*}) = Softmax(W_{y}^{(i)} \cdot X^{*} + b_{y}^{(i)}).
\end{split}
\end{equation}
Accordingly, the cross-entropy loss for the i-th FNN as:
 \begin{equation}
\begin{split}
\label{eq:8}
\mathcal{L}_{i} = - \frac{1}{P} \sum_{p  = 1}^{P} (y_{p}^{\top} \cdot log(\hat{y}_{p}^{(i)}) + (1 - y_{p})^{\top} \cdot log(1 - \hat{y}^{(i)}_{p})),
\end{split}
\end{equation}
where $P$ is the total number of patients.

Since the FFN ensemble contains up to $L$ FFNs, up to $L$ prediction results would be generated, corresponding to $L$ cross-entropy losses. Accordingly, the final prediction result is the sum of all members in the FFN ensemble. Subsequently, we apply an attention layer to $X^{*}$ to generate a series of attention weights as:
 \begin{equation}
\begin{split}
\label{eq:9}
\beta_{1}, \beta_{2}, \cdots, \beta_{L} = Softmax(W_{\beta} \cdot X^{*} + b_{\beta}),
\end{split}
\end{equation}
The contribution of those FNNs is gated by the generated attention weights as:
 \begin{equation}
\begin{split}
\label{eq:10}
\mathcal{L} = \sum_{i = 1}^{L} \beta_{i} \mathcal{L}_{i},
\end{split}
\end{equation}
where $\mathcal{L}$ is the final loss function.

\section{EXPERIMENTS}
\subsection{Datasets, Tasks, and Evaluation Metrics}
Based on the eICU$\footnote{https://eicu-crd.mit.edu/}$ Collaborative Research Database \cite{pollard2018eicu}, researchers have created benchmark datasets and proposed benchmarks/tasks \cite{sheikhalishahi2020benchmarking}. We focused specifically on predicting the in-hospital mortality risk of patients based on the data from the first 24 hours and 48 hours after eICU admission \cite{awad2017early}. The 27,390 patients/samples were extracted from the eICU database, where the Positive (likely to die)/Negative (unlikely to die) ratio is 1:6.15. We utilized sequential and non-sequential features as inputs for in-hospital mortality risk predictions \cite{ge2018interpretable}. The former refers to vital signs, while the latter refers to previous ICD-9 diagnosis codes. The vital sign measurements were gathered from the first 48 hours after eICU admission \cite{sheikhalishahi2020benchmarking}. Specifically, the 16 physiological variables were selected based on the literature \cite{sheikhalishahi2020benchmarking}. Table \ref{tab:physiological48hours} displays the results obtained from the preliminary analysis of physiological variables. 
The missing values of physiological variables were replaced with the empirical mean values \cite{che2018recurrent}. Detailed information on previous ICD-9 diagnosis codes can be found in the literature \cite{pollard2018eicu} (i.e., the DIAGNOSIS table).

Since the in-hospital mortality risk prediction problem can be viewed as a binary classification task, we assess the performance using AUROC, AUPRC, Precision, Recall, F1 Score, and Min(Se, P+) (i.e., the minimum of precision and sensitivity). The source code of our method and data extraction are released at the Github repository$\footnote{}$.

\begin{table}[htbp]
  \centering
  \caption{The 16 physiological variables selected from the eICU database.}
  \label{tab:intro2}
    \begin{tabular}{llr}
    \toprule
    Feature & Data Type  & \multicolumn{1}{l}{Missingness (\%)} \\
    \midrule
    Diastolic blood pressure & continuous & 33.80 \\
    Fraction inspired oxygen & continuous & 98.14 \\
    Glasgow coma scale eye & categorical & 83.42 \\
    Glasgow coma scale motor & categorical & 83.43 \\
    Glasgow coma scale total & categorical & 81.70 \\
    Glasgow coma scale verbal & categorical & 83.54 \\
    Glucose & continuous & 83.89 \\
    Heart Rate & continuous & 27.45 \\
    Height & continuous & 99.19 \\
    Mean arterial pressure & continuous & 96.53 \\
    Oxygen saturation & continuous & 38.12 \\
    Respiratory rate & continuous & 33.11 \\
    Systolic blood pressure & continuous & 33.80 \\
    Temperature & continuous & 76.35 \\
    Weight & continuous & 98.65 \\
    pH & continuous & 97.91 \\
    \bottomrule
    \end{tabular}%
  \label{tab:physiological48hours}%
\end{table}%

\subsection{Baselines}
We utilize seven deep learning models as baselines, including ICU-LSTM \cite{ge2018interpretable}, GRU \cite{cho2014learning}, GRU-D \cite{che2018recurrent}, Transformer \cite{vaswani2017attention}, AdaCare \cite{ma2020adacare}, Graph Attention Networks (GAT) \cite{velivckovic2017graph}, and Graph Convolutional Transformer (GCT) \cite{choi2020learning}. These recurrent neural network models, i.e., ICU-LSTM, GRU, and GRU-D, are described in the related work section. Transformer is the encoder of the Transformer \cite{vaswani2017attention}; the data is flattened and processed by FFNs to carry out in-hospital mortality risk predictions. AdaCare consists of a GRU, a multi-scale dilated convolution module, and a scale-adaptive clinical feature recalibration module. Both GAT and GCT are GNN-based methods. GAT is an early well-known approach that incorporates masked self-attention layers to improve the performance of graph convolution network-based methods. GCT is a combination of a Transformer and a Graph Convolutional Network. The Transformer captures the hidden EHR structure. The Graph Convolutional Network then learns the meaningful association between hidden feature representations. We also present a variant of our method (i.e., known as Our$_{\alpha}$), where hypergraph convolutional layers are replaced with graph convolutional layers.

\subsection{Implementations}
We randomly divide the data set into the training, validation, and testing sets in a 0.7:0.15:0.15 ratio. We train the proposed method using an Adam optimizer \cite{kingma2014adam} with a learning rate of 0.00039 and a mini-batch size of 256. The dimension of hidden states of GRU is 59. The number of layers of the Hypergraph Convolutional Network is 3. The value of threshold $\zeta$ is 0.4 (to be detailed in Section 6). The dimension size of $\Phi$ is 37. The number of FFNs in the FFN ensemble is 4. The number of layers of each FFN is 2, and the dimensions are 27 and 17. The dropout method is applied to the FFN ensemble, and the dropout rate is 0.2. All methods are implemented with PyTorch 1.10.0 on an Nvidia A40 GPU. All approaches are repeated ten times, and the average values with standard deviation for each evaluation metric are reported.

\section{Performance Evaluation}
Tables \ref{tab:prediction24hours} and \ref{tab:prediction48hours} (below) provide the results of in-hospital mortality risk prediction based on the data from the first 24 hours and 48 hours after eICU admission. Values in the parentheses are standard deviations. As can be seen from Tables \ref{tab:prediction24hours} and \ref{tab:prediction48hours}, the proposed method achieves superior performance over the baselines in terms of AUROC, AUPRC, Accuracy, Precision, F1 Score, and Min(Se, P+). The superior performance of Our than the Our$_{\alpha}$ verifies the efficacy of Hypergraph Convolutional Networks, which can capture the higher-order relationships among ICD-9 diagnosis codes and thus improve the prediction performance. The most interesting aspect of these two tables is that GRU-D is the best baseline and consistently outperforms other methods such as GRU, Transformer, and AdaCare.

Moreover, the prediction performance of all methods improved significantly as the prediction window from the first 24 hours to the first 48 hours after eICU admission. For instance, Transformer achieves an AUROC of 0.7869 based on the data from the first 48 hours after eICU admission, which is a significant improvement over 0.6881 based on the data from the first 24 hours after eICU admission. This result may be explained by the fact that due to the greedy nature of learning in deep neural networks, the performance of these prediction models largely depends on the size of the input data.

\section{Analysis of Hyper-parameter $\zeta$}
\begin{figure*}[!htbp]
        \centering
        \includegraphics[width = 1.0\linewidth]{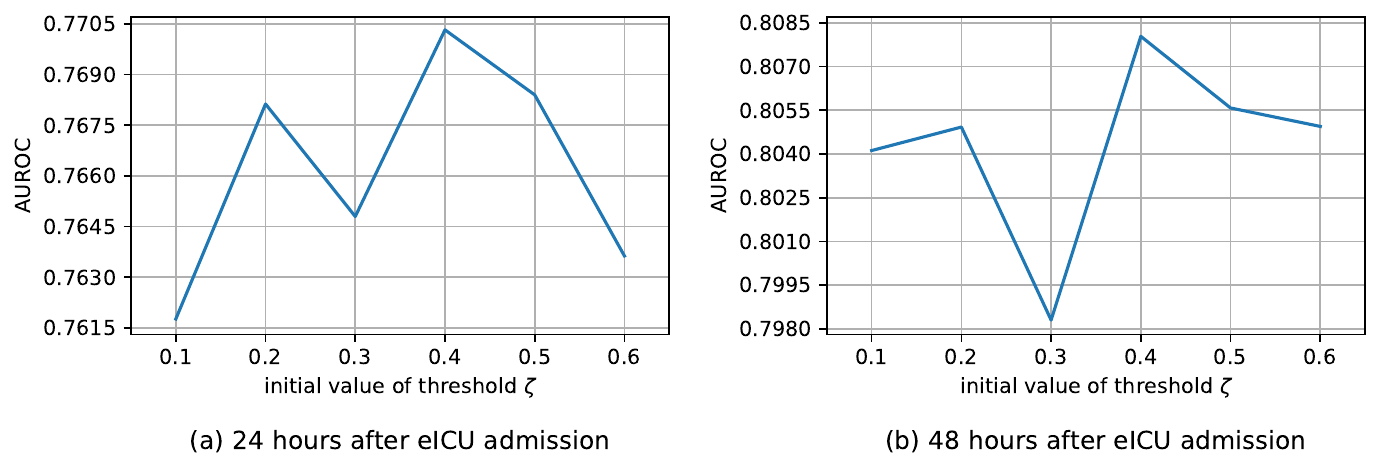}
        \caption{\textbf{The AUROC score of our method with different $\zeta$ values.}}
        \label{fig:Hyper-Parameter}
\end{figure*}

We analyze the prediction performance of our method with different $\zeta$ values. $\zeta$ is a hyper-parameter applied to the similarity computation among the patients. We tune hyper-parameter $\zeta$ between 0.1 and 0.6 to choose the best one. The AUROC score is used as a metric of the prediction performance, and the highest scores suggest the best performance of methods. From the graph below, we can see that different $\zeta$ values have a major influence on the model performance. Looking at Figure~\ref{fig:Hyper-Parameter} (below), it is apparent that the method with $\zeta$ of 0.4 reported significantly more AUROC scores than the others. Accordingly, we set the $\zeta$ to 0.4 for our model to avoid over-tuning this parameter for various datasets and tasks, as well as to reduce the computational complexity and time consumption.

\begin{table*}[htbp]
  \centering
  \caption{Performance of our method with other baselines on mortality risk prediction (24 hours after eICU admission).}
    \begin{tabular}{rclllllll}
    \toprule
    \multicolumn{2}{c}{Metrics} & \multicolumn{1}{c}{AUROC} & \multicolumn{1}{c}{AUPRC} & \multicolumn{1}{c}{Accuracy} \\
    \midrule
          & ICU-LSTM \cite{ge2018interpretable} & 0.7044(0.0001) & 0.2975(0.0009) & 0.6439(0.0074) \\
          & GRU \cite{cho2014learning} & 0.7006(0.0267) & 0.2847(0.0278) & 0.7292(0.0475) \\
          & GRU-D \cite{che2018recurrent} & 0.7571(0.0046) & 0.3456(0.0064) & 0.7032(0.0171) \\
    \multicolumn{1}{c}{\multirow{2}[0]{*}{Methods}} & Transformer \cite{vaswani2017attention} & 0.6881(0.0034) & 0.2867(0.0063) & 0.6757(0.0702) \\
          & AdaCare \cite{ma2020adacare} & 0.7465(0.0061) & 0.3180(0.0167) & 0.6926(0.0232) \\
          & GAT \cite{velivckovic2017graph} & 0.7228(0.0195) & 0.3153(0.0185) & 0.6607(0.0340) \\
          & GCT \cite{choi2020learning} & 0.7454(0.0060) & 0.3360(0.0097) & 0.6652(0.0221) \\
          & Our$_{\alpha}$ & 0.7437(0.0114) & 0.3282(0.0209) & 0.6612(0.1038) \\
          & Our & \textbf{0.7739(0.0027)} & \textbf{0.3588(0.0051)} & \textbf{0.7317(0.0092)} \\
    \midrule
    \multicolumn{2}{c}{Precision} & \multicolumn{1}{c}{Recall} & \multicolumn{1}{c}{F1 Score} & \multicolumn{1}{c}{Min(Se, P+)} \\
    \midrule
          & 0.2369(0.0030) & 0.6590(0.0074) & 0.3461(0.0024) & 0.3237(0.0021) \\
          & 0.2748(0.0227) & 0.5117(0.1324) & 0.3496(0.0251) & 0.3268(0.0263) \\
          & 0.2789(0.0095) & 0.6613(0.0230) & 0.3920(0.0064) & 0.3790(0.0046) \\
          & 0.2434(0.0217) & 0.5611(0.1138) & 0.3334(0.0051) & 0.3046(0.0066) \\
          & 0.2687(0.0318) & 0.6545(0.1190) & 0.3697(0.0293) & 0.3673(0.0327) \\
          & 0.2494(0.0174) & 0.6616(0.0520) & 0.3612(0.0173) & 0.3445(0.0255) \\
          & 0.2585(0.0092) & \textbf{0.7012(0.0353)} & 0.3773(0.0066) & 0.3661(0.0042) \\
          & 0.2625(0.0716) & 0.6605(0.1308) & 0.3790(0.0531) & 0.3454(0.0213) \\
          & \textbf{0.3026(0.0056)} & 0.6569(0.0170) & \textbf{0.4128(0.0063)} & \textbf{0.3929(0.0082)} \\
    \bottomrule
    \end{tabular}%
  \label{tab:prediction24hours}%
\end{table*}%

\begin{table*}[htbp]
  \centering
  \caption{Performance of our method with other baselines on mortality risk prediction (48 hours after eICU admission).}
    \begin{tabular}{rclllllll}
    \toprule
    \multicolumn{2}{c}{Metrics} & \multicolumn{1}{c}{AUROC} & \multicolumn{1}{c}{AUPRC} & \multicolumn{1}{c}{Accuracy} \\
    \midrule
          & ICU-LSTM \cite{ge2018interpretable} & 0.7373(0.0086) & 0.3354(0.0101) & 0.6525(0.0140) \\
          & GRU \cite{cho2014learning} & 0.7558(0.0221) & 0.3368(0.0243) & 0.7434(0.0709) \\
          & GRU-D \cite{che2018recurrent} & 0.7925(0.0066) & 0.3845(0.0103) & 0.7401(0.0205) \\
    \multicolumn{1}{c}{\multirow{2}[0]{*}{Methods}} & Transformer \cite{vaswani2017attention} & 0.7869(0.0030) & 0.3971(0.0198) & 0.7202(0.0306) \\
          & AdaCare \cite{ma2020adacare} & 0.7832(0.0053) & 0.3724(0.0078) & 0.6950(0.0433) \\
          & GAT \cite{velivckovic2017graph} & 0.7838(0.0091) & 0.3823(0.0133) & 0.6926(0.0416) \\
          & GCT \cite{choi2020learning} & 0.7898(0.0052) & 0.3915(0.0105) & 0.6998(0.0108) \\
          & Our$_{\alpha}$ & 0.7950(0.0236) & 0.3996(0.0260) & 0.7538(0.0643) \\
          & Our & \textbf{0.8141(0.0096)} & \textbf{0.4109(0.0156)} & \textbf{0.7579(0.0133)} \\
    \midrule
    \multicolumn{2}{c}{Precision} & \multicolumn{1}{c}{Recall} & \multicolumn{1}{c}{F1 Score} & \multicolumn{1}{c}{Min(Se, P+)} \\
    \midrule
          & 0.2578(0.0079) & 0.6965(0.0163) & 0.3762(0.0083) & 0.3604(0.0110) \\
          & 0.3201(0.0442) & 0.5228(0.2603) & 0.3484(0.1200) & 0.3737(0.0215) \\
          & 0.3330(0.0158) & 0.7189(0.0422) & 0.4544(0.0122) & 0.4189(0.0070) \\
          & 0.3122(0.0190) & 0.7008(0.0670) & 0.4299(0.0061) & 0.3986(0.0082) \\
          & 0.2985(0.0236) & 0.7414(0.0580) & 0.4238(0.0163) & 0.4011(0.0094) \\
          & 0.2960(0.0211) & 0.7417(0.0426) & 0.4220(0.0174) & 0.3998(0.0135) \\
          & 0.3017(0.0049) & \textbf{0.7571(0.0299)} & 0.4313(0.0046) & 0.4079(0.0083) \\
          & 0.3412(0.0427) & 0.7008(0.1527) & 0.4426(0.0424) & 0.4081(0.0241) \\
          & \textbf{0.3503(0.0089)} & 0.7143(0.0342) & \textbf{0.4689(0.0105)} & \textbf{0.4322(0.0127)} \\
    \bottomrule
    \end{tabular}%
  \label{tab:prediction48hours}%
\end{table*}%

\section{Model Effectiveness and Robustness}
Now we focus on predicting the mortality risk in patients with a particular disease, i.e., Congestive Heart Failure (CHF) \cite{han2022early} and Diabetes \cite{anand2018predicting}. Data were gathered from patients with CHF and Diabetes based on the first 24 hours and 48 hours after eICU admission. For the data from the first 24 hours after eICU admission, the sample size of CHF patients is 2953, where the Negative (unlikely to die)/Positive (likely to die) ratio is 4.9898:1; the sample size of Diabetes patients is 840, where the Negative/Positive ratio is 9.1204:1. For the data from the first 48 hours after eICU admission, the sample size of CHF patients is 2818, where the Negative/Positive ratio is 5.1934:1; the sample size of Diabetes patients is 780, where the Negative/Positive ratio is 10.3043:1.

Tables \ref{tab:Personalized1} and \ref{tab:Personalized2} show the results obtained from the in-hospital mortality risk prediction in CHF and Diabetic patients. Group I represents the data from patients with a particular disease, while Group II represents all the remaining data. Overall, Group II reported significantly more AUROC, AUPRC, Precision, Recall, F1 Score, and Min(Se, P+) scores than Group I. It seems possible that these results are due to differences in the size and quality of the datasets.

Furthermore, there are significant differences between the prediction outcomes in CHF and Diabetic patients. For instance, from the data in Table \ref{tab:Personalized1}, Group I in CHF patients reported significantly more AUROC, AUPRC, Precision, F1 Score, and Min(Se, P+) scores than Group I in Diabetic patients. These results demonstrate the effectiveness and robustness of our proposed network construction, leveraging the foundation of Hypergraph Convolutional Networks (i.e., based on prior medical knowledge) to contribute to personalized mortality risk prediction. A further study with more focus on diseases with high mortality, such as sepsis \cite{su2022early} and acute kidney injury \cite{lin2019predicting} in ICU, is therefore suggested.

\begin{table*}[htbp]
  \centering
  \caption{Performance of our method on mortality risk prediction in CHF and Diabetic patients (24 hours after eICU admission).}
    \begin{tabular}{p{10em}lllllll}
    \toprule
    \multicolumn{1}{c} {Training Set} & \multicolumn{1}{c} {AUROC} & \multicolumn{1}{c} {AUPRC} & \multicolumn{1}{c} {Accuracy} \\
    \midrule
    Group I: CHF & 0.6770(0.0139) & 0.2587(0.0114) & 0.6908(0.0985) \\
    Group II: w/o CHF & 0.7482(0.0099) & 0.3284(0.0147) & 0.6891(0.0759) \\
    Group I: diabetic & 0.6083(0.0277) & 0.2144(0.0223) & 0.3888(0.1913) \\
    Group II: w/o diabetes & 0.7583(0.0085) & 0.3430(0.0137) & 0.6770(0.0440) \\
    \midrule
    \multicolumn{1}{c} {Precision} & \multicolumn{1}{c} {Recall} & \multicolumn{1}{c} {F1 Score} & \multicolumn{1}{c} {Min(Se, P+)} \\
    \midrule
    0.2527(0.0383) & 0.5062(0.1760) & 0.3172(0.0287) & 0.2975(0.0128) \\
    0.2766(0.0310) & 0.6607(0.1269) & 0.3827(0.0191) & 0.3594(0.0141) \\
    0.1679(0.0229) & 0.7614(0.2261) & 0.2643(0.0234) & 0.2561(0.0328) \\
    0.2701(0.0190) & 0.7091(0.0579) & 0.3896(0.0144) & 0.3775(0.0165) \\
    \bottomrule
    \end{tabular}%
  \label{tab:Personalized1}%
\end{table*}%

\begin{table*}[htbp]
  \centering
  \caption{Performance of our method on mortality risk prediction in CHF and Diabetic patients (48 hours after eICU admission).}
    \begin{tabular}{p{10em}lllllll}
    \toprule
    \multicolumn{1}{c} {Training Set} & \multicolumn{1}{c} {AUROC} & \multicolumn{1}{c} {AUPRC} & \multicolumn{1}{c} {Accuracy} \\
    \midrule
    Group I: CHF & 0.7339(0.0105) & 0.3217(0.0103) & 0.6611(0.0987) \\
    Group II: w/o CHF & 0.7892(0.0094) & 0.3768(0.0131) & 0.7239(0.0302) \\
    Group I: diabetic & 0.6309(0.0378) & 0.2393(0.0322) & 0.3915(0.2007) \\
    Group II: w/o diabetes & 0.7939(0.0064) & 0.3851(0.0091) & 0.7171(0.0333) \\
    \midrule
    \multicolumn{1}{c} {Precision} & \multicolumn{1}{c} {Recall} & \multicolumn{1}{c} {F1 Score} & \multicolumn{1}{c} {Min(Se, P+)} \\
    \midrule
    0.2718(0.0431) & 0.6609(0.1436) & 0.3728(0.0236) & 0.3612(0.0105) \\
    0.3158(0.0179) & 0.7025(0.0726) & 0.4336(0.0124) & 0.4020(0.0092) \\
    0.1821(0.0360) & 0.7822(0.2241) & 0.2810(0.0266) & 0.2775(0.0438) \\
    0.3130(0.0207) & 0.7205(0.0577) & 0.4346(0.0126) & 0.4094(0.0095) \\
    \bottomrule
    \end{tabular}%
  \label{tab:Personalized2}%
\end{table*}%

\section{Conclusion}
In ICU settings, patients often require continuous monitoring to ensure timely care by healthcare professionals. Thus, developing and introducing machine learning-based models to assist healthcare professionals in predictive clinical and managerial decision-making is needed. In this paper, we present a Hypergraph Convolutional Network for fine-grained ICU patient similarity analysis and risk prediction. To the best of our knowledge, this is the first study that uses a tailored Hypergraph Convolutional Network for similarity computation among patients in ICU settings. Our proposed Hypergraph Convolutional Network emphasized the benefits of constructing hypergraphs of prior medical knowledge for downstream clinical predictions. Experiments manifest that our proposed method has higher AUROC, AUPRC, Precision, F1 Score, and Min(Se, P+) against state-of-the-art deep prediction models on the publicly available eICU Collaborative Research Database. Furthermore, the effectiveness and robustness of the model decisions are demonstrated via several case studies.

\section{Limitations and Future Works}
The present study was limited in several ways. (i) Although the proposed method has achieved superior performance in in-hospital mortality risk predictions, there is abundant room for further optimizing the network architecture. In future investigations, it might be possible to reduce the complexity of neural network architecture in which the hyper-parameters and hidden layers as well as the number of nodes can be tuned. (ii) The proposed method may not be applicable to all learning settings. For example, this includes transfer learning, few-shot learning, and zero-shot learning required for real application scenarios. Considerably more work will need to be done to incorporate the proposed method into the above learning settings. (iii) An issue not addressed in this study was the temporality of physiological variables, i.e., varying time intervals between physiological variables. Further studies regarding the effect of the temporality of physiological variables on the risk prediction task would be worthwhile. (iv) The most important issue is that the proposed method remains "black-box" in terms of its architecture. Therefore, the learning process of the model is non-transparent, which leads to difficulties in results interpretation. Further research is required to establish the transparency and explainability of the proposed method. (v) Further research could also be conducted to determine the effectiveness of the proposed method on the MIMIC-III \cite{johnson2016mimic} and MIMIC-IV \cite{johnson2023mimic} databases.

\bibliographystyle{splncs04}
\bibliography{BIB}
\end{document}